# Eclectic Extraction of Propositional Rules from Neural Networks

Ridwan Al Iqbal
Department of Computer Science
American International University-Bangladesh
Banani, Dhaka, Bangladesh
ridwan@enosisbd.com

*Abstract* – **Artificial Neural Network is among the most popular algorithm for supervised learning. However, Neural Networks have a well-known drawback of being a "Black Box" learner that is not comprehensible to the Users. This lack of transparency makes it unsuitable for many high risk tasks such as medical diagnosis that requires a rational justification for making a decision. Rule Extraction methods attempt to curb this limitation by extracting comprehensible rules from a trained Network. Many such extraction algorithms have been developed over the years with their respective strengths and weaknesses. They have been broadly categorized into three types based on their approach to use internal model of the Network. Eclectic Methods are hybrid algorithms that combine the other approaches to attain more performance. In this paper, we present an Eclectic method called HERETIC. Our algorithm uses Inductive Decision Tree learning combined with information of the neural network structure for extracting logical rules. Experiments and theoretical analysis show HERETIC to be better in terms of speed and performance.**

## I. INTRODUCTION

Artificial Neural Networks have been one of the most popular and successful supervised learning methods. It is being widely used successfully in many practical domains. The reason of such popularity is its ability to recognize any function and complex non-linear relationships [1] [2]. It is also robust to noise in data and it is capable of online learning. Moreover, Neural Networks are also more suitable to handle real valued data. So, Neural Networks have shown to be a more accurate classifier in many domains compared to symbolic methods such as C4.5.

Nevertheless, despite their predictive ability, ANN and related discriminative models such as Support Vector Machines have well-known drawbacks; such as their black-box approach to modeling and the resultant lack of transparency. What can be learnt from their systemic underlying knowledge representation is little more than a set of weights, activation functions and optimal parameters, discovered during the Neural Network training. It is not easily comprehensible to Human users what knowledge these weights actually represent. This makes it difficult to gain credibility for such learners. Despite attaining higher accuracy, symbolic learning algorithms are preferred in many application domains such as Business Intelligence, Automated Diagnosis and so on [3].

Therefore, obtaining a meaningful interpretation of trained Neural Networks is a significant improvement; as it would allow the strengths of neural networks such as noise robustness and accuracy to be available to high fidelity learning tasks. Significant research has been done over the years to compensate this inadequacy in Neural Networks. Over 30 different algorithms have been proposed since the 90's [4]. These algorithms have shown different performance characteristics and have their respective strengths and weaknesses.

This paper proposes a novel rule extraction algorithm called HERETIC (Hierarchical and Eclectic Rule Extraction via Tree Induction and Combination). HERETIC uses Decision Tree Induction in a novel way to induce symbolic rules from trained Feed forward Neural Networks or Multilayer Perceptrons (MLP). HERETIC is computationally fast and also more accurate than existent rule extraction methods.

The paper is organized as follows; Section 2 gives an overview of previous rule extraction literature; Section 3 gives background information on Neural Networks and also the motivation behind HERETIC; Section 4 describes the HERETIC algorithm; Section 5 describes the experimentation with the algorithm on some real world dataset and performance analysis.

We now describe the notation used in this paper. N is the number of instances while K is the number of features. The output from a neural network is *y(X)* given an instance x. Consequently, y also represents the actual output node whenever mentioned in a subscript. The components of neural networks are represented in a layered fashion. A superscript always represents the source layer while subscripts represent nodes between successive layers. Therefore, $w_{ij}^k$ is the weight from node i of layer k to node j of layer k+1. In the fashion, $h_j^n$ is the output of a node j in layer n. $\theta_j$ is the bias of node j. $\delta_j$ represents the error of unit *j*; $\alpha$ is the learning rate. *Layer$_k$* or *L$_k$* is the set of units of layer k, where *k* is 1 for the first hidden layer; *Inputs(k)* is the set of all units that is input to *k*. $\mu$ represents mean. *U* is the set of all network units.

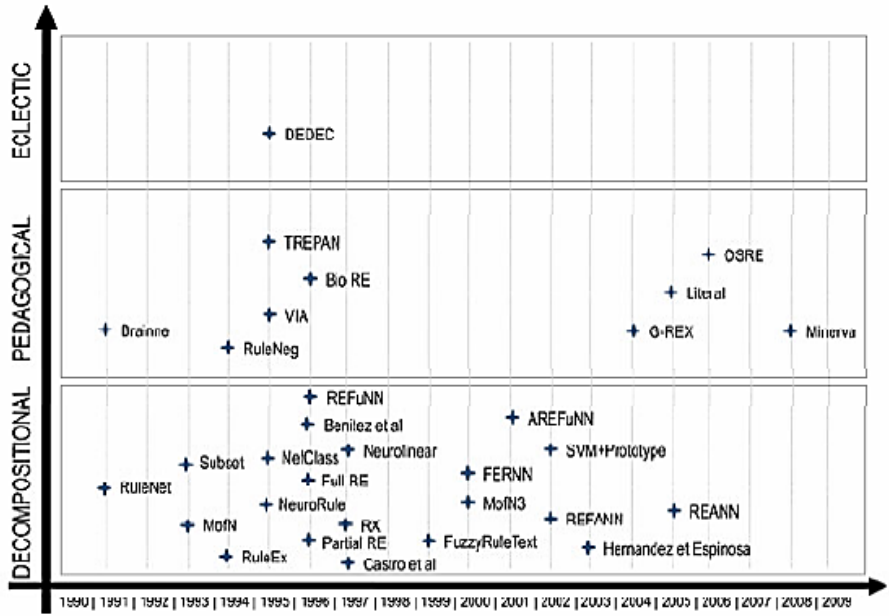

Fig. 1: Timeline of various Rule Extraction algorithms

## II. PAST RESEARCH REVIEW

### A. Rule Extraction Taxonomy

Rule Extraction from Neural Networks has seen extensive research in the past 2 decades. There are some surveys that extensively explore this research area [5] [6]. Andrews, Diederich and Tickle developed an overall taxonomy for categorizing techniques that extract rules from trained Neural Networks [7]. The taxonomy recommends five primary criteria.

1) The representation language of the extracted rules; 2) the translucency of the view of the underlying network architecture; 3) the quality of extracted rules; 4) the computational complexity of the rule extraction technique; and 5) the portability across multiple architectures.

Of all the criterion, translucency criteria is the most important as it defines the approach taken by an algorithm. This criterion defines whether an algorithm has knowledge about the internal structure of the network.

Under classification criterion (2), at one end of the spectrum we have those rule extraction techniques that use most information from the network structure and view the underlying ANN at the maximum level of granularity i.e. as a set of discrete hidden and output units. Craven & Shavlik [8] categorized such techniques as "Decompositional." The basic motif of Decompositional rule extraction techniques is to extract rules at the level of each individual hidden and output unit. They also use some heuristic to analyze the weights of the connections to confer the rules.

In contrast to the Decompositional approaches, the theme in the Pedagogical approaches is to view the trained ANN as a "black box". That is it can be only used as an oracle but its internal structure will not be exposed. The focus is then on finding rules that map the inputs into outputs using some form of symbolic training procedure. So, Pedagogical methods perform symbolic supervised learning to find rules.

Eclectic methods combine the previous approaches. They analyze the ANN at the individual unit level but also extract rules via training instead of analyzing weights. Figure 1 shows a time line of various rule extraction algorithms.

We now describe the past literature on these approaches.

### B. Decompositional Methods

Most decompositional methods search for a combination of weights that would cause activation to a particular unit. The SUBSET method by Towell and Shavlik is an example of such methods. The MofN algorithm [9], is an extension of SUBSET. It clusters the weights of a trained network into equivalence classes and extracts m-of-n style rules. The algorithm was later extended into other decompositional algorithms such as MofN3 by Setino [10], a new method for extracting M-of-N rules from neural networks. There are several problems with subset style algorithms. These methods require a hard limiting threshold function or an approximated logistic activation function for hidden units to extract rules. The search space is exponential thus limiting effectiveness in large networks. Many later researchers tried to limit the search space by using heuristics [10].

RULEX algorithm works in a different fashion. It performs rule extraction by directly converting weight vectors to rules rather than searching for subsets. The RULEX technique is exclusively developed to extract rules from the constrained error Backpropagation CEBPL network that performs function approximation and classification very similar to radial basis function networks [11]. In the recent years, R. Setiono, W. K. Leow and Jack M. Zurada [12] described a method called rule extraction from function approximating neural networks (REFANN) and Functional extraction of Neural Networks (FERNN) for extracting rules from trained neural networks for nonlinear regression. But it only supports networks with only one hidden layer.

Some Decompositional algorithms have shown very good accuracy. But as most algorithms either rely on analyzing weights or searching for subsets from weights, they suffer from the exponential worst case time. Moreover, Decompositional methods are not portable across different Network types.

*C. Pedagogical Methods*

One of the earliest Pedagogical methods was developed by Saito and Nakano [13]. The idea was to search for combinations of input values which activate each output unit. One problem with this method is that the size of the search space can grow exponentially with the number of input values. The authors used two heuristics that limited the search space.

Gallant [14] developed a method similar to Saito and Nakano's method. Gallant's method uses a procedure to test the combinations of input values (rules) against the network. Thrun [15] developed a method called *Validity Internal Analysis* (VIA). In this method, linear programming is used to determine if a set of constraints placed on the network activation values is consistent.

The Trepan algorithm by Craven [16], extracts a decision tree from a trained network. The trained network is used as an "oracle" that is able to answer queries during the learning process. The oracle (the network) determines the class of each instance that is presented in the query.

OSRE is one of the recent pedagogical methods [17]. OSRE extends the algorithm proposed by Ruleneg [18] to ordinal and continuous variables using trained data to perform a 1-from-N coding, while searching, in orthogonal directions, where the decision surface crosses a decision boundary.

The main advantage of Pedagogical methods is that they are not dependent on the internal structure of neural networks. So, they are highly portable to many different types of network architectures, even any other arbitrary learner such as SVM. However, the main issue with pedagogical methods is performance. Because they are highly portable, they cannot take full advantage of the learned Neural Network structure. The training required is also quite extensive. The training essentially means having to relearn an already learned model using symbolic learning algorithms. So, the advantage of using neural networks as an intermediate stage is minimal.

*D. Eclectic Methods*

Eclectic methods combine the previous approaches. They analyze the Neural Network architecture at the individual unit level but also extract rules using training. One example of this approach is the method proposed by Tickle et al. called DEDEC [19]. It is applicable to a broad class of multilayer feed forward ANNs trained by the Backpropagation algorithm. This method works in two steps. It identifies functional dependencies between inputs and outputs of an ANN by analyzing the architecture and weight vectors of the trained network. Then the next phase is essentially pedagogical as symbolic learning is performed based on the weight analysis.

Eclectic methods have not been well explored as the other two mainstream approaches. But there are several gains from such a hybrid approach. Using the knowledge of the architecture would enable us to use as much information as possible from the network. Moreover, the symbolic learning used by pedagogical methods is more robust to learn patterns than simple search for subset.

III. BACKGROUND AND MOTIVATION

*A. Neural Networks*

A Neural network is composed of several neurons. Each neuron is a processing element with a linear model. In a feed forward network, the network architecture composes a hierarchy of nodes where output of lower layer becomes input of higher layer. Each unit represents a simple linear combination of weighted input which is then goes through an activation function to generate output [20].

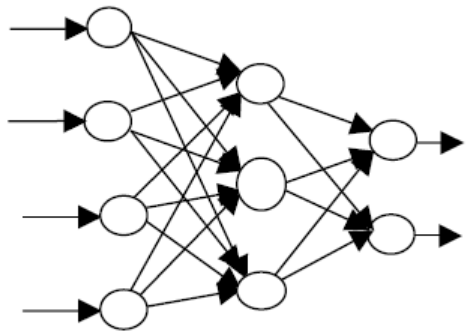

Fig. 2: A sample feed forward neural network.

The simplest form is a perceptron. Which is a single neuron processing inputs. The equation of a perceptron is given by.

$$f(\mathcal{X}) = \sum w_i x_i + \theta$$

$$y(\mathcal{X}) = \sigma(f(\mathcal{X})) \qquad (1)$$

Here, $f(\mathcal{X})$ is the linear combination of input vectors. It is then passed onto the activation function $\sigma$ to generate the final output. The activation function generally takes the form of a sigmoid function. The logistic function is the most popular choice.

$$\sigma(x) = \frac{1}{1+e^{-x}}$$

A single neuron is only capable of solving problems that are linearly separable [20]. Based on this formulation of a perceptron, other neural network architectures are formed. As shown in figure 2, a feed forward neural network or Multilayer perceptron is composed of hierarchical layers of neurons. Each neuron is capable of learning a simple concept. But each successive layer is a composition of simpler concept which forms a more complex pattern [21].

The following shows the output function of a multilayer perceptron with n layers [20].

$$y(X) = \sigma \left[ \sum_{j \in L_n} w_{jy}^n h_j^n + \theta_y \right] \quad (2)$$

$$h_j^k = \sigma \left[ \sum_{i \in L_{k-1}} w_{ij}^{k-1} h_i^{k-1} + \theta_j \right]$$

$$h_j^0 = X_j$$

*B. Motivation Behind HERETIC*

Each neuron represents a simple pattern in a Neural Network. So, the problem of extracting rules can be simplified if the extraction is done at each neuron level instead of the whole network. This is what the Decompositional methods do.

Moreover, some Pedagogical methods use symbolic learning methods. For example Trepan as mentioned earlier uses Neural Network as an oracle to generate new training set to build a decision tree. However, it can be argued that training a neural network from the training set is not beneficial in this case as ultimately another training phase needs to be performed to generate the tree which can be done with the training set alone [11]. Training a decision tree from training data is a hard problem in itself, training a decision tree from a neural network is even harder. So this poses question as to why train from a network that is itself not fully accurate.

The idea behind HERETIC is to use the advantage of both methods. Decomposition into individual units would simplify the problem of extraction; while symbolic training is better than searching for subset of weights. We would perform symbolic training at each neuron. So, each Node in ANN would generate a decision tree trained via the input output set generated from each neuron. The output of the trees of the previous layer would then become the input of the trees of the next layer. This hierarchical structure of Trees approximates the Neural Network.

In terms of symbolic training, any method that learns a set of rules from a training set would suffice. We can use Inductive Rule Learning methods such as FOIL [22] and RIPPER [23] as well. However, decision tree algorithms have been shown to be very fast and accurate rule learner and they are widely used. There have been many modification and additions into the basic Decision Tree Induction method and its performance has been widely studied. That's why we have chosen a decision tree learning algorithm such as C4.5 by Quinlan [24]. After the Trees are generated, they can be easily converted into rules of Disjunctive Normal Form.

## VI. DESCRIPTION OF HERETIC

*A. The Algorithm*

The basic operation of HERETIC is to construct a Decision Tree at every node of the Neural Network. One of the key advantages of HERETIC is that it supports different types of network architectures. The neural network can have any number of layers and the Network can also be partially connected. In fact, it can be adopted for recurrent networks to some extent. This strength comes from the fact that a Decision Tree is a universal approximator [25]. It can learn any function. As each unit in a Neural Network learns a function, it can also be approximated by a Decision Tree.

However, for understandability of the rules generated, we have restricted activation function of the ANN to be only sigmoid type. The first step of Rule extraction is to train the Neural Network with the training set. Then the network input outputs are discretized. Any discretization algorithm can be used for this purpose. We assume each unit can have only two possible output values 0 and 1. So, we are restricting the units to be binary units. This assumption can cause a loss of precision as the output of a Neural Network Unit ranges from 0 to 1 when logistic activation is used.

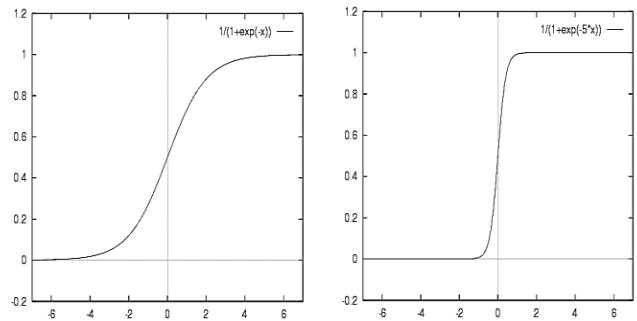

Fig. 3: Approximating a step function with logistic function. The left figure shows the normal logistic curve. The right curve shows a modified logistic curve.

However, we can easily resolve this by using a steep sigmoid function that approximates a step function. Multiplying a large constant m with the input would make the function steeper. As neural network takes a normalized input from 0 to 1, so the constant should be a large value and it is recommended to use a value greater than 100. This enables the logistic function to behave as a step function yet being differentiable. Therefore,

$$\sigma(x) = \frac{1}{1+e^{-mx}} \qquad m > 100$$

The derivative of the logistic function would change as well. The new derivative would be,

$$\frac{d}{dx}\sigma(x) = \frac{d}{dx}\frac{1}{1+e^{-mx}}$$

$$= m\,\sigma(x)(1-\sigma(x))$$

Thus, this would add this constant k into the learning rule of Backpropagation. So, the new rule for Backpropagation will be,

$$\Delta w_{ij}^k = m\,\alpha\,X_{ij}^k \sum_{p\,\epsilon\,L_{k+1}} \delta_p^{k+1} w_{jp}^{k+1} \qquad (3)$$

$$\delta_d = (t_d - y_d)y_d(1-y_d)h_j^N$$

Neural Network trained with this rule would be used for rule extraction by HERETIC.

After training and discretization, the next step is to actually generate the Tree using C4.5. Each unit generates a tree so each unit would need its own training set from which to learn. The training set given to learn the ANN can be used for this purpose. The samples from training set are fed into the trained network. Each sample generates output in the Neurons. Output of the previous layer becomes the input of the next layer. This way we can find input output pair for all Neurons. The resultant training sets would then be used to train the Decision Trees. As Each Decision Tree Learning is independent from others, it is also possible to learn them in parallel.

The Decision Tree learning algorithm used is at the discretion of the users. Popular and older Learning Methods like C4.5 or ID3 can be used. More recent enhancements to the basic algorithm can also be used as long as the output is a univariate tree.

After tree generation, each tree is converted into a rule of disjunctive normal form. This can be easily done as each unique path in a tree represents a conjunction of features that classifies instances. So, each unique path in the tree is a Decision rule.

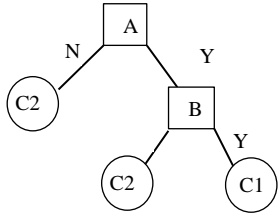

Fig. 4: A sample Decision Tree

This sample tree has three unique paths that can be converted into the following rules.

*A=Y & B=Y then C1*

*A=N then C2*

*A=Y & B=N then C1*

Then we can convert these conjunctive rules into one disjunctive rule per class.

*(A=Y & B=Y) OR (A=Y & B=N) then C1*

*A=N then C2*

This way each decision tree can be converted into two disjunctive rules as our Neuron units have only two output value. Finally, we generate one rule per class for the whole Neural Network via substitution. As each unit of the higher layer has lower layer units as input, therefore we can replace the lower layer input symbol with its respective rule.

For example, a sample ruleset of a output layer is shown below.

Output unit rules:

*X=1 OR (Y=0 & Z =1) then Class =1*

*X=0 then Class =0*

Hidden unit X rules:

*(A=1 & B=0) OR (A=0 & C=1) then X =1*

*(A=0 & B=0) OR (A=1 & C=0) then X =0*

Substituting X with its rules:

*((A=1 & B=0) OR (A=0 & C=1)) OR (Y=0 & Z =1) then Class =1*

*(A=0 & B=0) OR (A=1 & C=0) then Class =0*

This way we can recursively substitute until the original inputs remain. Certainly such a rule would be incomprehensible as the ANN itself. So, the final step of our algorithm is the logical simplification. There are many algorithms available for this purpose [26]. Most algorithms are based on heuristics and run in quadratic time in the number of variables. Any such minimizer can be used. We used the popular espresso logic minimization algorithm [26]. The final output provided will be simplified rules in Disjunctive Normal Form.

**HERETIC Algorithm**

***Input:*** *Training Set **T***

***Output:*** *Ruleset **R***

1. Train the Neural Network **ANN** with modified sigmoid function using **S**.
2. **For each** samples in S,
   Generate output from the trained **ANN.** Generate separate training set $S_i$ for each neuron $N_i$.
3. **For Each** training set $S_i$
   Generate Decision Tree $T_i$ using **C4.5**.
4. **Convert** each Tree into Disjunctive rules
5. **Substitute** recursively from the output unit rules to the original feature set.
6. **Simplify** using Logical Minimizer **Espresso algorithm.**

*B. Computational complexity*

The time complexity for HERETIC is dependent upon the steps performed by the algorithm. Training the neural network is not part of the complexity as HERETIC only extracts rules and it is possible to provide a trained neural network as input. Given training set size n, Number of features f , number of neurons u, number of connections w, number of epochs e; we assume the network to be fully connected. The time complexity of training a neural network would be: $O(enw)$. $w$ can at most be $u^2$ in the worst case. So time for training a neural network:

$$T_{ANN} = O(enu^2)$$

Generating input output for each neuron would require only $O(nw)$ time. Generating a tree using C4.5 requires $O(nk^2)$ for k number of features [24]. In this particular case, the number of features would be dependent upon the number of neurons in the previous layer. Assume $h_k$ to be number of units of layer k with. So, for units of layer k, generating tree would require $O(np_{k-1}^2)$. Thus, time complexity for generating trees,

$$T_G = nf^2h_1 + nh_1^2h_2 + nh_2^2h_3 \ldots \ldots nh_{k-1}^2h_k$$

$$T_G > nh_{max}^2 u\,;$$

$h_{max}$ = the layer with maximum number of units

$$T_G = O(nh_{max}^2 u) \quad (4)$$

This learning process takes the most time as heuristic logical minimization algorithms take quadratic time in the number of variables [26]. So time for logical minimization is $O(f^2)$.

So, the total time complexity of HERETIC,

$$T = O(nh_{max}^2 u) \quad (5)$$

This is much better than training the neural network as $h_{max} \leq w$ and actually much lower than w in most cases. Number of units $u$ is normally small in practical applications. Generating all decision trees would be also be much faster for the fact that ANN is an iterative method and attaining convergence is normally slow requiring very high number of epochs.

This is also true for many rule extractions methods that employ searching for weight combination. Time for HERETIC is an enormous reduction to the decompositional methods, which take exponential time in the worst case [7].

## V. EXPERIMENTAL STUDIES

### A. Experimental Setup

We tested our algorithm with several benchmark datasets from the University of California, Irvine (UCI) dataset repository. We compare the accuracy of the extracted rules in terms of test data. We also study the Fidelity of the rules. Fidelity refers to how closely the rule classification matches the original neural network. The performance of our algorithm is compared with two other rule extraction techniques, Trepan and FERNN. These were chosen as a representative of their respective categories. We chose Pedagogical method Trepan and Decompositional method FERNN. We initially set the neural network into only 1 hidden layer. After several iterations with different configurations, the best performing network architecture was chosen. These datasets were trained with at most 2 hidden layers. Nominal features were converted to several binary variables. Real features were normalized.

The standard Backpropagation algorithm was used with weight decay.

Table I: Description of datasets

| Dataset Name | Samples | Features | ANN nodes |
|---|---|---|---|
| Promoters | 936 | 58 | 20 h1 |
| Breast Cancer | 286 | 9 | 11 h1, 3 h2 |
| Congress Vote | 435 | 16 | 6 h1 |
| Heart Disease | 303 | 75 | 14 h1, 6h2 |
| Monks 1 | 432 | 8 | 10 h1 |
| Monks 2 | 432 | 8 | 10 h1 |
| Monks 3 | 432 | 8 | 11 h1 |

For the Monks problems, separate training and test set is provided so we tested with that split. 10 fold cross validation was performed on other datasets. The training was continued for 200 epochs. The learning rate chosen was very small to compensate for the large constant k that is multiplied due to the modified training rule. We chose a learning rate of 0.002 and k value of 100 which brings the overall learning rate to 0.2 which is a reasonable learning rate.

We used the Weka implementation for both C4.5 and ANN. Standard entropy based split function was used along with pruning. The pruning dataset was kept 20% of the training set. The whole experiment was conducted 20 times with the mean value taken.

### B. Results and Performance analysis

The following Table II & III shows the performance of algorithms compared to HERETIC. The experimental results show that HERETIC performs very well in converting a trained ANN into rules. The performance of HERETIC is close to the source ANN. This is shown by the fidelity for the various datasets. Fidelity is more than 90% in all datasets. In fact, Paired-T statistical test shows no significant difference in all datasets apart from Heart Disease and Monks 2 dataset. Moreover, HERETIC has accuracy similar or better than the other two algorithms in some cases. However, it performs best in terms of fidelity. It was better than both other rule extraction algorithms in terms of fidelity. It bested Trepan in all datasets and FERNN in some datasets. Trepan performs the least in all the datasets, proving our analysis made earlier that learning only one decision tree from a neural network is a hard problem with few real performance gains.

The performance of C4.5 was quite good compared to ANN but still ANN was slightly better in most cases. It shows that instead of extracting rules from ANN it may often be easier to simply use Decision Tree to quickly generate rules. But ANN along with rule extraction performed best in all the datasets; and there were significant difference in performance in most datasets between C4.5 and ANN based algorithms.

TABLE II: ACCURACY OF VARIOUS ALGORITHMS ON DIFFERENT DATASETS

| Methods | Promoters | | Breast Cancer | | Heart Desease | | Vote | | Monks 1 | | Monks 2 | | Monks 3 | |
|---|---|---|---|---|---|---|---|---|---|---|---|---|---|---|
| | Accu. | StdDev | Accu. | StdDev | Accu. | StdDev | Accu. | StdDev | Accu. | StdDev | Accu. | StdDev | Accu. | StdDe |
| **ANN** | 94.83 | 1.52 | 95.57 | 0.511 | 87.83 | 0.606 | 96.35 | 1.546 | 100.0 | 0.672 | 98.84 | 1.453 | 100 | 0.309 |
| **HERETIC** | 94.75 | 0.12 | 95.87 | 0.293 | 85.33 | 0.934 | 96.54 | 0.382 | 100.0 | 0.341 | 97.45 | 0.724 | 99.43 | 0.763 |
| Trepan | 85.42 | 1.63 | 91.23 | 1.864 | 78.42 | 0.471 | 90.19 | 0.564 | 98.34 | 0.034 | 87.94 | 2.289 | 95.39 | 0.871 |
| FERNN | 91.72 | 0.54 | 95.81 | 0.55 | 82.23 | 0.591 | 94.70 | 1.366 | 97.50 | 0.253 | 94.95 | 0.348 | 98.98 | 0.65 |
| C4.5 | 85.64 | 0.72 | 88.78 | 2.581 | 79.64 | 0.28 | 93.42 | 0.122 | 100.0 | 0.483 | 75.34 | 0.233 | 98.98 | 0.15 |

TABLE III: COMPARSION OF FIDELITY SCORES BETWEEN THE RULE EXTRACTION ALGORITHMS

| Methods | Promoters | | Breast Cancer | | Heart Desease | | Vote | | Monks 1 | | Monks 2 | | Monks 3 | |
|---|---|---|---|---|---|---|---|---|---|---|---|---|---|---|
| | Fidel. | StdDev | Fidel. | StdDev | Fideli. | StdDev | Fidel. | StdDev | Fideli. | StdDev | Fidel. | StdDev | Fidel. | StdDe |
| **HERETIC** | 99.43 | 0.289 | 98.35 | 0.902 | 99.10 | 0.934 | 99.34 | 0.382 | 100.0 | 0.843 | 98.74 | 0.488 | 99.19 | 0.463 |
| Trepan | 94.22 | 0.348 | 94.56 | 0.063 | 89.34 | 0.471 | 95.09 | 0.564 | 99.34 | 0.918 | 91.46 | 0.243 | 93.23 | 0.341 |
| FERNN | 87.56 | 0.233 | 97.84 | 0.479 | 93.45 | 0.591 | 97.90 | 1.366 | 98.45 | 0.469 | 95.32 | 0.492 | 99.59 | 0.359 |

## V. CONCLUSION AND FUTURE DIRECTIONS

We proposed a fast and highly accurate rule extraction algorithm HERETIC. The theoretical and experimental results show that using an eclectic approach that combines Neural Network Architecture with symbolic training is a considerably fast and accurate way to extract meaningful rules from neural network. This algorithm will be useful for many different Network Architecture that many Decompositional methods do not support while being very accurate. Use of logical minimization is another novel approach applied by HERETIC which ensures compact rule set that will be more comprehensible to users. Experiments also show that use of this combined approach has huge performance gain then simply using a symbolic learning algorithm. So, for many learning problems where a meaningful and understandable learned model is needed, our approach would bring better performance along with good understandability.

This research can be further enhanced by extending HERETIC for generating rules of First Order Logic or Description logics. Moreover, the assumption that Neuron output would only be binary is resolved by a special training rule. This is another good research problem as to if HERETIC could be extended for network units that support smoother activation functions and also functions other than sigmoid such as RBF.